\makeatletter\def\input@path{{}{sty/}}\makeatother
\documentclass[10pt,twocolumn,letterpaper]{article}

\usepackage[pagenumbers]{cvpr}   %

\usepackage{graphicx}
\usepackage{amsmath}
\usepackage{amssymb}
\usepackage{booktabs}
\usepackage{multirow}
\usepackage{xcolor}
\usepackage{caption}
\usepackage{cuted}        %
\usepackage{currfile}
\expandafter\def\csname ver@everyshi.sty\endcsname{2001/05/15}
\usepackage{tikz}         %
\usetikzlibrary{arrows.meta,positioning,calc,shapes.geometric}

\newif\ifdraft\drafttrue

\ifdraft

  \newcommand{\TODO}[1]{{\color{red}\textbf{[TODO: #1]}}}
\else

  \newcommand{\TODO}[1]{}
\fi

\newcommand{\fmr}[1]{FMR=$10^{-#1}$}
\newcommand{\FMR}{FMR}
\newcommand{\FNMR}{FNMR}
\newcommand{\Neff}{N_{\mathrm{eff}}}
\newcommand{\cond}[1]{\textsc{#1}}   %

\draftfalse

\definecolor{cvprblue}{rgb}{0.21,0.49,0.74}
\usepackage[pagebackref,breaklinks,colorlinks,allcolors=cvprblue]{hyperref}
\hypersetup{
  pdfauthor={Fu Dai, Kamal Taha, Naoufel Werghi, Maregu Assefa},
  pdftitle={Contamination at the Operating Point},
  pdfsubject={},
  pdfkeywords={},
  pdfcreator={},
  pdfproducer={},
}
\pdftrailerid{}

\newcommand{\paperID}{}
\newcommand{\confName}{}
\newcommand{\confYear}{}

\title{Contamination at the Operating Point:\\
       Train--Test Identity Overlap Inflates Low-\FMR{} Face Recognition
       Accuracy by an Order of Magnitude}

\author{Fu Dai \quad Kamal Taha \quad Naoufel Werghi \quad Maregu Assefa\\}

\begin{document}
\maketitle

\begin{abstract}
Face verification systems are deployed by selecting a decision threshold
that achieves a target false match rate (\FMR{}); performance is then
reported as the corresponding false non-match rate (\FNMR{}). In practice,
deployed systems operate at \FMR{} $\le 10^{-6}$, but widely used benchmarks
can share identities with the models' training corpora. Prior work measures
the resulting \emph{optimistic bias} as a small additive
change in 10-fold accuracy (up to $1.01$ percentage
points)~\cite{wu2024overlap}. We show that the bias is neither small nor constant: it is a multiplicative error that
grows as \FMR{} falls. First, we audit train-test overlap with a detector calibrated against a null distribution (rather than a hand-tuned threshold). Relative to WebFace42M, we
flag $88.6\%$ of MS1MV3 identities and $14.3\%$ of IJB-C subjects as
present, despite WebFace260M's published de-overlap explicitly naming
IJB-C. Manual review of $100$ flagged IJB-C subjects confirms $69.2\%$,
implying a residual overlap of roughly $9.9\%$.
Second, holding the model fixed, removing only the contaminated identities from the
test set increases \FNMR{} by $3.6\times$ at \fmr{4}, $6.3\times$ at
\fmr{5}, and $13.9\times$ at \fmr{6}; while a same-size random removal does not. The effect is traced to a shorter impostor tail that lowers the threshold. The magnitude of the shift depends on the training
corpus and does not cancel between models, reducing the apparent margin
between our model and a public baseline by half. Third, we give the first direct measurement of the effective
number of independent impostor comparisons in IJB-C: only $2.7$ million effectively independent comparisons at \fmr{6}, which is below all commonly
accepted reporting rules. Finally, by fixing a public model with a known training corpus, we build a real-face protocol with audited
zero overlap that resolves \fmr{8}.
\end{abstract}

\section{Introduction}
\label{sec:intro}

\begin{figure}[t]
\centering
\includegraphics[width=\linewidth]{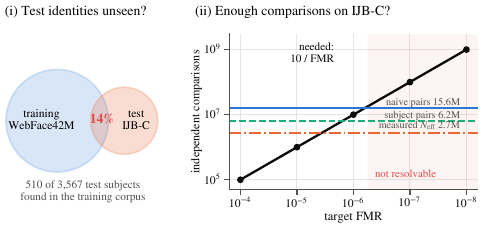}
\caption{Motivation. (i)~After the published de-overlap, $14\%$ of IJB-C
subjects are still found in the WebFace42M training corpus. (ii)~IJB-C has
too few independent impostor comparisons to resolve \fmr{6} or lower
(needed: $10/\FMR{}$, black).}
\label{fig:motivation}
\end{figure}

\begin{figure*}[t]
\centering
\includegraphics[width=\linewidth]{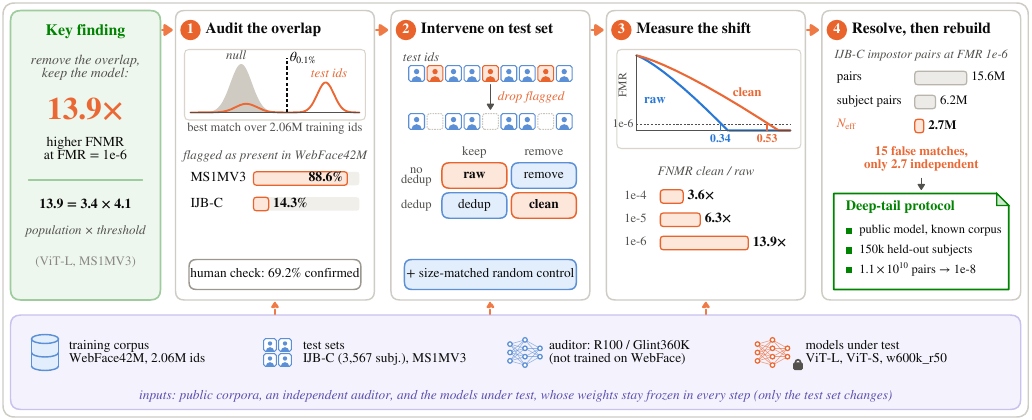}
\caption{Overview of the framework. Left: the headline result for ViT-L on
MS1MV3, factored into a population and a threshold component
(\cref{sec:bias:mechanism}). (1)~A null-calibrated audit flags test
identities present in WebFace42M (cut $\theta_{0.1\%}$: $0.1\%$ false
positives on a masked-self null); a human check confirms $69.2\%$ of the
IJB-C flags. (2)~Flagged identities are removed from the \emph{test} set
only, in a $2\times2$ design with deduplication and a size-matched random
control. (3)~Overlap shortens the impostor tail and lowers the threshold at
each \FMR{}, so \FNMR{} inflation grows from $3.6\times$ at \fmr{4} to
$13.9\times$ at \fmr{6} (tail curves schematic; thresholds measured).
(4)~IJB-C has only $2.7$M effectively independent impostor comparisons at
\fmr{6}, so we build a real-face protocol with audited zero overlap that
resolves \fmr{8}. Bottom: the inputs; model weights never change.}
\label{fig:framework}
\end{figure*}

A face \emph{verification} system compares two face images and returns a
similarity score (here, the cosine similarity between two embeddings).
Comparing two images of the same person is a \emph{genuine} (mated)
comparison; comparing images of two different people is an \emph{impostor}
(non-mated) comparison. For a decision threshold $\tau$, the \emph{false
match rate} $\FMR(\tau)$ is the fraction of impostor comparisons scored at
or above $\tau$, and the \emph{false non-match rate} $\FNMR(\tau)$ is the
fraction of genuine comparisons scored below $\tau$~\cite{zhu2023webface260m_pami,iso19795}.
Systems are reported at an \emph{operating point}: $\tau$ is
chosen so that \FMR{} equals a target such as $10^{-6}$, and performance is
summarised by the corresponding \FNMR{}, written \FNMR{}@\fmr{6}
(equivalently TAR@FAR, its complement).
Public benchmarks such as IJB-C are most often reported at
\fmr{4}~\cite{zhu2023webface260m_pami,dias2026ms1mv3}, whereas NIST
reports its visa and border tracks at \fmr{6}~\cite{zhu2023webface260m_pami},
and one-to-many search against large galleries may require \fmr{8} or lower.

Every such figure rests on two assumptions: (i) the test identities do not
appear in the training data, and (ii) the test protocol contains enough
\emph{independent} comparisons to resolve the reported operating point.
This paper shows that both assumptions fail for widely used benchmarks
(\cref{fig:motivation}).

\paragraph{Overlap is known; its impact at the operating point is not.}
We refer to the error caused by training on test identities as
\emph{optimistic bias}: the difference between the \FNMR{} a model achieves
on a benchmark that shares identities with its training corpus and the
\FNMR{} it achieves, at the same \FMR{}, on unseen identities.
Dulhanty and Wong~\cite{dulhanty2020auditing} report a $4.0$-point gap in
closed-set rank-1 accuracy on VGGFace2, and Wu~\etal~\cite{wu2024overlap}
find that $46.93\%$ of LFW identities appear in MS1MV2 and measure a bias of
$0.04$ to $1.01$ percentage points in 10-fold verification accuracy. Neither
study reports \FNMR{} or TAR@FAR. The question \emph{``by how much is the published
\FNMR{} at \fmr{6} wrong?''} remains open.

\paragraph{Why the answer depends on the operating point.}
Margin-based losses~\cite{deng2019arcface} are optimised to separate the
\emph{training} identities. On seen identities, the largest
impostor scores are suppressed, so the
inferred threshold is biased downward, and more so deeper in the tail
(\cref{sec:bias}). Holding the model fixed and
removing only the train-overlapping identities from the test set raises
\FNMR{} by $3.6\times$, $6.3\times$, and $13.9\times$ at \fmr{4}, \fmr{5},
and \fmr{6}, respectively.

\paragraph{Auditing overlap requires a calibrated cut.}
WebFace260M~\cite{zhu2021webface260m,zhu2023webface260m_pami} flags a
test identity as overlapping when its best match exceeds a
\emph{hand-set threshold} (cosine $0.7$), but the best-match score among
millions of candidates depends on the embedding space and on the search
size. We instead set the cut at the $99.9$th percentile of a null of
best-match scores for faces known \emph{not} to be in the training set, so
the audit's false-positive rate is $0.1\%$ by construction
(\cref{sec:method:detector}), and use a matcher trained on an independent
corpus (Glint360K). With this detector, WebFace260M's
published de-overlap still leaves $510$ of IJB-C's subjects ($14.3\%$)
flagged as present in WebFace42M, with $148$ above the authors' own $0.7$
cut. Manual verification of $100$ flagged subjects confirms $69.2\%$,
implying residual overlap near $9.9\%$.

\paragraph{Is the deep tail measurable at all?}
An \FMR{} can only be resolved if enough independent false matches are
expected: Doddington's \emph{rule of 30}~\cite{doddington2000nist} calls
for at least $30$ errors, and the looser \emph{rule of ten}, common in
computer vision, for $10$. How many comparisons are independent depends on
the counting convention (\cref{sec:method:resolution}): the \emph{naive}
count of impostor pairs, the \emph{subject-pair} count $\binom{G}{2}$ for
$G$ subjects, or the \emph{pessimistic} count $\min(n,t)$ of Mansfield and
Wayman~\cite{mansfield2002best}. Under IJB-C's official
protocol~\cite{maze2018ijbc}, these are $15{,}638{,}932$,
$\binom{3531}{2}=6{,}232{,}215$, and $3{,}531$, so under the rule of ten
\fmr{6} is just resolvable, just not resolvable, or not resolvable by three
orders of magnitude; \cref{sec:neff} measures which holds.

\paragraph{Framework and contributions.}
\cref{fig:framework} summarises the analysis. Our contributions are:
\begin{enumerate}[leftmargin=*]\itemsep0pt
\item We show that optimistic bias from train--test identity overlap is a
      function of the operating point, increasing by roughly an order of
      magnitude from \fmr{4} to \fmr{6} on MS1MV3, and we measure the
      threshold-displacement mechanism behind it.
\item We introduce a null-calibrated overlap audit that avoids hand-set
      thresholds and show that WebFace260M's de-overlap leaves roughly
      one in seven IJB-C subjects flagged as present.
\item We provide the first direct measurement of the effective number of
      independent impostor comparisons in IJB-C: $2.7$ million at
      \fmr{6}, too few to report \fmr{6}.
\item By fixing a public model with a known training corpus, we build a
      real-face protocol that resolves \fmr{8} with audited zero overlap.
\end{enumerate}

\section{Related Work}
\label{sec:related}

\subsection{Training and evaluation datasets}
\label{sec:related:datasets}

Face recognition models are typically trained on large, web-scraped collections
organised by (putative) celebrity identity and evaluated on smaller benchmarks;
\cref{tab:datasets} lists those we refer to. WebFace42M is distilled from
WebFace260M by automatic self-training (CAST), and its name list is partly
inherited from MS1M: ``the MS1M identities are most a subset of the WebFace
ones''~\cite{zhu2023webface260m_pami}.

\begin{table}[t]
\centering\footnotesize
\setlength{\tabcolsep}{3pt}
\resizebox{\linewidth}{!}{%
\begin{tabular}{lrrll}
\toprule
Dataset & IDs & Images & Role & Protocol \\
\midrule
CASIA-WebFace~\cite{yi2014casia}      & $10$K  & $0.5$M & train & -- \\
VGGFace2~\cite{cao2018vggface2}       & $9$K   & $3.3$M & train & -- \\
MS-Celeb-1M~\cite{guo2016msceleb}     & $100$K & $10$M  & train & ${\sim}50\%$ noise \\
MS1MV2~\cite{deng2019arcface}         & $85$K  & $5.8$M & train & -- \\
MS1MV3~\cite{deng2019lightweight}     & $93{,}431$ & $5{,}179{,}510$ & train & -- \\
Glint360K~\cite{an2021glint360k}      & $360$K & $17$M  & train & -- \\
WebFace260M~\cite{zhu2021webface260m} & $4$M   & $260$M & train & -- \\
WebFace42M~\cite{zhu2023webface260m_pami} & $2{,}059{,}906$ & $42{,}474{,}558$ & train & -- \\
\midrule
LFW~\cite{huang2008lfw}               & $5{,}749$ & $13{,}233$ & test & 10-fold acc. \\
CFP-FP~\cite{sengupta2016cfp}, AgeDB-30~\cite{moschoglou2017agedb} & -- & -- & test & 10-fold acc. \\
IJB-C~\cite{maze2018ijbc}             & $3{,}531$ & -- & test & TAR@FAR $\ge10^{-6}$ \\
WebFace260M test & $2{,}478$ & $57{,}715$ & test & \FNMR{}@\fmr{5} \\
DigiFace-1M~\cite{bae2023digiface}    & $99{,}999$ & -- & test & synthetic \\
\bottomrule
\end{tabular}}
\caption{Training and evaluation datasets as reported
in~\cite{zhu2023webface260m_pami,wu2024overlap,dias2026ms1mv3}.}
\label{tab:datasets}
\end{table}

\subsection{Error rates and how many errors an estimate needs}
\label{sec:related:metrics}

Verification performance is commonly reported either as 10-fold pair-classification
accuracy on the LFW family~\cite{wu2024overlap} or as an operating point on the
\FMR{}/\FNMR{} trade-off described in \cref{sec:intro}. The latter follows the
ISO/IEC 19795-1 convention~\cite{iso19795}, which WebFace260M adopts~\cite{zhu2023webface260m_pami}.
Doddington~\etal~\cite{doddington2000nist} discuss the ``rule of 30'' for NIST speaker
evaluations (\cref{sec:intro}); the looser ``rule of ten'' corresponds to $10$ errors
and an uncertainty of approximately $\pm62\%$.
Mansfield and Wayman~\cite{mansfield2002best} give a pessimistic variance estimate
$V(\FMR)=\alpha(1-\alpha)/\min(n,t)$, which later summaries of the ISO/IEC 19795 practice also cite~\cite{thieme2015standards,iyer2015rules};
Schuckers~\cite{schuckers2010computational} presents a parametric treatment.
Fogliato~\etal~\cite{fogliato2024ci} derive the asymptotic variance of an \FMR{}
estimate for 1:1 matching, with an effective sample size lower-bounded by $G/2$ for
$G$ subjects (\cref{sec:method:neff}).
Fallahi~\etal~\cite{fallahi2025reliability} apply reliability criteria across $62$
biometric benchmarks and derive a minimum reportable \FMR{} of $6\times10^{-5}$ for
IJB-C, noting that subject count is not directly incorporated.
In \cref{sec:neff} we quantify, on a real benchmark, where the effective count lies
between $\binom{G}{2}$ and $G/2$.

\subsection{Cleaning a corpus and choosing its thresholds}
\label{sec:related:cleaning}

Large-scale training corpora are typically cleaned using an embedding-based face
matcher with hand-chosen cosine-similarity thresholds.
WebFace260M's CAST pipeline~\cite{zhu2023webface260m_pami} keeps the largest DBSCAN
cluster in each name folder, merges or deletes folders whose feature centres are
similar (merge above cosine $0.7$, delete the smaller between $0.5$ and $0.7$), and
removes overlap with popular benchmarks ``if the cosine similarity is higher than
$0.7$''~\cite{zhu2023webface260m_pami}.
Deng~\etal~\cite{deng2019arcface} remove MegaFace/FaceScrub overlap from training
without quantifying the residual impact.
Dias~\etal~\cite{dias2026ms1mv3} audit MS1MV3 for \emph{intra-set} defects with a
centroid threshold and perceptual hashing, and compare each targeted removal with a
size-matched random-removal control; we follow this design.
Wu~\etal~\cite{wu2024overlap} match each LFW image to its nearest MS1MV2 images using
an independent Glint360K-trained ArcFace-R100 and annotate every candidate pair by hand.
Our detector (\cref{sec:method:detector}) adopts their independent matcher but replaces
hand-tuned thresholds with a cut calibrated under a null hypothesis.

\subsection{Current public model performance}
\label{sec:related:sota}

\cref{tab:sota} lists published IJB-C results. On their own test set,
Zhu~\etal{} report \FNMR{}@\fmr{5} of $2.98\%$ (WebFace42M) versus $9.88\%$
(MS1MV2), and a ResNet-200 entry with \FNMR{}@\fmr{6} of $0.0034$ (visa) and
$0.0088$ (border) on NIST-FRVT~\cite{zhu2023webface260m_pami}. On the LFW family,
six margin losses retrained on MS1MV2 span only $0.17$ points in average
accuracy~\cite{wu2024overlap}, less than the optimistic bias measured by Wu~\etal{},
and the deepest commonly reported public operating point remains \fmr{6} on IJB-C.

\begin{table}[t]
\centering\footnotesize
\setlength{\tabcolsep}{3pt}
\resizebox{\linewidth}{!}{%
\begin{tabular}{llrrrrl}
\toprule
Backbone & Training data & $10^{-6}$ & $10^{-5}$ & $10^{-4}$ & $10^{-3}$ & Source \\
\midrule
ResNet-100 & WebFace42M & -- & $96.23$ & $97.70$ & -- & \cite{zhu2023webface260m_pami} \\
ResNet-100 & MS1MV2     & -- & $94.05$ & $96.03$ & -- & \cite{zhu2023webface260m_pami} \\
IResNet-50 & MS1MV3     & $90.39$ & $94.68$ & $96.39$ & $97.53$ & \cite{dias2026ms1mv3} \\
ResNet-100 & Glint360K  & -- & -- & $97.55$ & -- & model zoo \\
ResNet-50  & WebFace600K & -- & -- & $97.25$ & -- & model zoo \\
\bottomrule
\end{tabular}}
\caption{Published IJB-C TAR (\%) at FAR $10^{-6}$ to $10^{-3}$. The last two rows are the public models we evaluate
(\cref{sec:setup:models}).}
\label{tab:sota}
\end{table}

\subsection{Train--test overlap in face recognition}
\label{sec:related:overlap}

Dulhanty and Wong~\cite{dulhanty2020auditing} partition VGGFace2 probes via name
matching against an MS1M-RetinaFace training list and report closed-set rank-1 of
$79.71\%$ for in-list identities versus $75.73\%$ for out-of-list identities.
Wu~\etal~\cite{wu2024overlap} is the closest work to ours. After identifying $2{,}009$
LFW identities in MS1MV2 (\cref{sec:intro}), they construct three training subsets
(ID-Disjoint, ID-Overlap-R, and ID-Overlap-C), retrain six methods on each, and
measure an optimistic bias across five LFW-family benchmarks (mean $0.44$), with larger
effects on harder sets.
They intervene on the \emph{training} set and measure a causal effect; we intervene on
the \emph{test} set under a fixed model and measure reporting bias along the operating point.
\"Ozt\"urk and Marcel~\cite{ozturk2026membership} show, via membership inference, that
margin-based losses shape the embedding geometry of training identities differently
from unseen ones; we measure the implications of this phenomenon for threshold
selection.
Larson~\etal~\cite{larson2026rvlcdip} show that removing test--train duplicates from
RVL-CDIP decreases reported document-classification accuracy.
Concurrently, Khazem~\cite{khazem2026identities} starts from the same premise as
\cref{sec:neff} and derives a distribution-free false-accept guarantee rather than an
effective sample size.

\section{Methodology}
\label{sec:method}

\subsection{Scores, thresholds, and the two error rates}
\label{sec:method:scores}

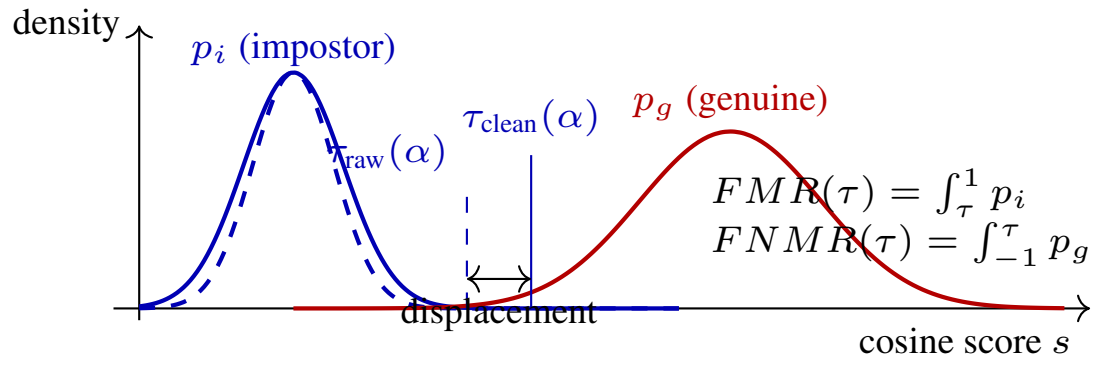
\begin{figure}[t]
\centering
\begin{tikzpicture}[x=0.85cm,y=0.78cm, font=\scriptsize]
  \draw[->] (-0.2,0) -- (7.4,0) node[below left] {cosine score $s$};
  \draw[->] (0,-0.1) -- (0,2.4) node[left] {density};
  \draw[thick, color=blue!70!black] plot[domain=0:4.2, samples=80, smooth]
      (\x, {2.0*exp(-((\x-1.2)/0.55)^2)});
  \draw[thick, dashed, color=blue!70!black] plot[domain=0:4.2, samples=80, smooth]
      (\x, {2.0*exp(-((\x-1.2)/0.45)^2)});
  \draw[thick, color=red!70!black] plot[domain=1.2:7.2, samples=80, smooth]
      (\x, {1.5*exp(-((\x-4.6)/1.0)^2)});
  \draw[blue!70!black] (3.05,0) -- (3.05,1.3) node[above] {$\tau_{\text{clean}}(\alpha)$};
  \draw[blue!70!black, dashed] (2.55,0) -- (2.55,1.0) node[above left] {$\tau_{\text{raw}}(\alpha)$};
  \node[blue!70!black] at (1.2,2.2) {$p_i$ (impostor)};
  \node[red!70!black] at (4.6,1.75) {$p_g$ (genuine)};
  \node[align=left, anchor=west] at (4.3,0.75) {$\FMR(\tau)=\int_\tau^1 p_i$ \\ $\FNMR(\tau)=\int_{-1}^{\tau} p_g$};
  \draw[<->, thin] (2.55,0.25) -- (3.05,0.25) node[midway, below] {displacement};
\end{tikzpicture}
\caption{Schematic illustration of Score densities and the operating point (measured
tails in \cref{sec:bias:mechanism}). $\tau(\alpha)$ leaves a fraction
$\alpha$ of the impostor density $p_i$ above it; \FNMR{} is the genuine
mass $p_g$ left of it. Test identities seen in training compress the
extreme impostor tail (dashed), so $\tau(\alpha)$ sits further left and the
\FNMR{} read at it is smaller.}
\label{fig:densities}
\end{figure}

Let $f(x)\in\mathbb{R}^d$ denote the $\ell_2$-normalised embedding of a face image $x$
under a fixed model, and let $s(x_a,x_b)=f(x_a)^{\top}f(x_b)$ be the cosine similarity
between two images, so $s\in[-1,1]$. Over a test set, genuine comparisons induce a score density
$p_g(s)$, while impostor comparisons induce a score density $p_i(s)$. For a threshold
$\tau$, we define
\begin{equation}
\FMR(\tau)=\int_{\tau}^{1} p_i(s)\,ds, \qquad
\FNMR(\tau)=\int_{-1}^{\tau} p_g(s)\,ds .
\label{eq:fmr_fnmr}
\end{equation}
An operating point fixes a target $\alpha$ and reports the corresponding \FNMR{} at the
threshold implied by the impostor distribution (e.g., \FNMR{}@\fmr{6} when
$\alpha=10^{-6}$):
\begin{equation}
\begin{aligned}
\tau(\alpha) &= \min\{\tau : \FMR(\tau)\le\alpha\},\\
\FNMR\text{@}\alpha &= \FNMR(\tau(\alpha)).
\end{aligned}
\label{eq:operating_point}
\end{equation}
In practice, $\tau(\alpha)$ is the $\lceil \alpha N\rceil$-th largest of the $N$ impostor
scores. Thus, at $\alpha=10^{-6}$ the threshold is determined by only the most confusable
impostor pairs---precisely the extreme tail that margin-based objectives are designed to
push apart. Test-set contamination suppresses this extreme tail of $p_i$ without materially
shifting the bulk distribution, moving $\tau(\alpha)$ left and yielding a smaller \FNMR{}
when evaluated on $p_g$ (\cref{fig:densities}).

\subsection{What \FMR{} can a test set resolve?}
\label{sec:method:resolution}

Consider an operating point with \FMR{} $\alpha$ computed over $N$ (nominal) non-mated
comparisons. The expected number of false matches supporting the estimate is
$k=N\alpha$. Reliability heuristics require $k\ge30$ under the rule of
30~\cite{doddington2000nist} or $k\ge10$ under the rule of ten (\cref{sec:intro}).
Accordingly, a benchmark with $N$ comparisons has a \emph{resolvable-\FMR{} floor}
\begin{equation}
\alpha_{\min}=k_{\min}/N,\qquad k_{\min}\in\{10,30\},
\label{eq:floor}
\end{equation}
below which it should not report an \FMR{} value.

The appropriate choice of $N$ depends on the counting convention (\cref{sec:intro}).
Between the naive pair count $N_{\text{pair}}$ (over images or templates) and the
pessimistic effective count $N_{\text{pess}}=\min(n,t)$~\cite{mansfield2002best},
a subject-pair view bounds the number of independent non-mated comparisons that $G$
subjects can support by
\begin{equation}
N_{\text{subj}}=\binom{G}{2}=\tfrac{1}{2}G(G-1),
\label{eq:subject_pairs}
\end{equation}
regardless of how many image pairs those subjects expand into.

The smallest $G$ satisfying $\binom{G}{2}\ge k/\alpha$ is therefore the minimum subject
count required by an operating point. Under the rule of ten, this yields $448$,
$1{,}415$, $4{,}473$, $14{,}143$, and $44{,}722$ subjects for \fmr{4} through \fmr{8};
under Doddington's rule of 30~\cite{doddington2000nist} the corresponding requirements
are $776$, $2{,}450$, $7{,}747$, $24{,}496$, and $77{,}461$ (the vertical lines in
\cref{fig:capability}). Under the pessimistic $\min(n,t)$ convention, the subject
requirement coincides with the pair requirement.

IJB-C has $3{,}531$ subjects, which falls short of the $4{,}473$ required for
\fmr{6}. The WebFace260M test set (\cref{tab:datasets}) naively appears to resolve
\fmr{8}, but its $\binom{2478}{2}=3{,}069{,}003$ subject pairs give $30.7$ expected
false matches at \fmr{5} and $3.1$ at \fmr{6}; its authors report \FNMR{} at \fmr{5}.

In our results, we report the number of independent subject pairs alongside each operating
point and mark points with fewer than ten expected false matches as \emph{under-resolved}.

\subsection{A null-calibrated overlap detector}
\label{sec:method:detector}

A fixed cosine threshold, such as WebFace260M's $0.7$ (\cref{sec:related:cleaning}),
does not determine the audit's false-positive rate: the maximum similarity over millions of
candidates can be high even when the probe identity is absent, and depends on the model,
image quality, and search size. We therefore calibrate the cut from a null distribution.

(i)~\emph{Representatives}: we keep one embedding per training identity ($2{,}059{,}907$ for
WebFace42M, one more label than reported in the paper).
(ii)~\emph{Probes}: for each test identity, we take up to $c$ images, score each against
\emph{all} training representatives, and retain the maximum
$m_j=\max_{x\in\text{id}_j,\;r}\, s(x,r)$.
(iii)~\emph{Null}: we repeat the same search for $4{,}000$ training representatives while masking
out each representative's own entry. The resulting distribution of $m$ corresponds to the
``best match among two million candidates'' for a face that is \emph{definitely not} in the set,
under the same model and search size.

We declare a test identity present when $m_j$ exceeds the $(1-\beta)$ quantile of the null,
$\theta_\beta$. We report $\beta=0.1\%$ and release the full sweep; by construction, the
false-positive rate at any cut is therefore known.

Crucially, the auditor must not be trained on the corpus it audits, since it would score its
own identities highly regardless. Following Wu~\etal~\cite{wu2024overlap}, all audit figures use
a public Glint360K-trained ArcFace-R100 (\cref{sec:setup:models}), and WebFace-trained models
appear only as agreement arms (\cref{sec:audit}).

\begin{figure}[t]
\centering
\begin{tikzpicture}[font=\scriptsize, scale=0.86, every node/.style={transform shape}]
  \begin{scope}
    \fill[blue!15] (0,0) circle (1.25);
    \fill[red!15] (1.5,0) circle (0.8);
    \begin{scope}
      \clip (0,0) circle (1.25);
      \fill[purple!45] (1.5,0) circle (0.8);
    \end{scope}
    \draw[blue!60!black, thick] (0,0) circle (1.25);
    \draw[red!60!black, thick] (1.5,0) circle (0.8);
    \node[blue!60!black] at (-0.45,0.3) {training};
    \node[blue!60!black] at (-0.45,-0.05) {identities};
    \node[red!60!black] at (1.85,0.25) {test};
    \node[purple!70!black, align=center] at (1.0,-0.55) {overlap};
    \node[align=center, text width=3.3cm] at (0.75,-1.75) {(a) \textbf{De-overlap}: remove from the \emph{test} set every identity flagged as present in the training set (model fixed).};
  \end{scope}
  \begin{scope}[xshift=4.6cm]
    \draw[red!60!black, thick] (0,0) circle (1.25);
    \node[red!60!black] at (0,0.95) {test identities};
    \fill[gray!60] (-0.55,0.15) circle (0.17); \fill[gray!60] (-0.25,0.15) circle (0.17);
    \draw[thick] (-0.55,0.15) -- (-0.25,0.15);
    \fill[gray!60] (0.45,-0.35) circle (0.17); \fill[gray!60] (0.75,-0.35) circle (0.17); \fill[gray!60] (0.6,-0.65) circle (0.17);
    \draw[thick] (0.45,-0.35) -- (0.75,-0.35) -- (0.6,-0.65) -- cycle;
    \fill[gray!25] (-0.7,-0.6) circle (0.17); \fill[gray!25] (0.2,0.45) circle (0.17); \fill[gray!25] (-0.1,-0.4) circle (0.17);
    \node[align=center, text width=3.3cm] at (0,-1.75) {(b) \textbf{Self-deduplication}: link test identities whose cosine exceeds $T$ and collapse each linked component to one subject.};
  \end{scope}
\end{tikzpicture}
\caption{The two cleaning operations. (a) De-overlap between corpora: the
purple region is what the audit flags and \cond{clean} removes. (b)
Deduplication within the test set: the same person under two labels would
otherwise produce an ``impostor'' score near $1$ and set the deepest
thresholds (\cref{sec:protocol}). Both act on the test set only; the model
is never changed.}
\label{fig:circles}
\end{figure}
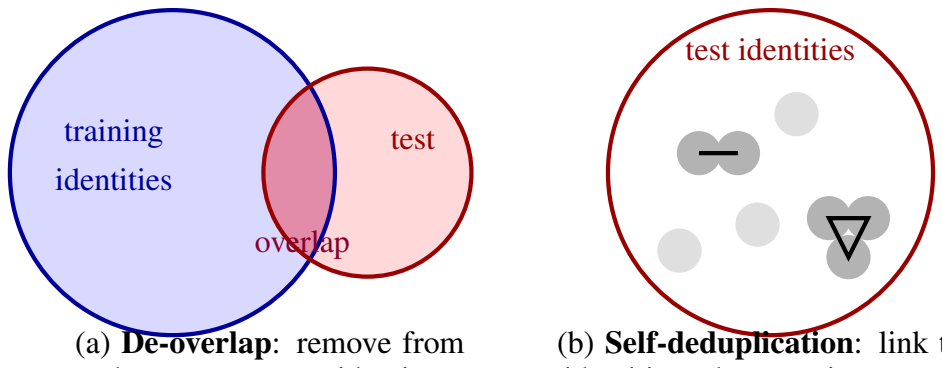

\subsection{Cleaning as a precision--recall trade-off}
\label{sec:method:cleaning}

Both operations of \cref{fig:circles} are classifiers over identity pairs,
with \emph{recall} the fraction of true overlaps (or duplicates) removed and
\emph{precision} the fraction of removals that were true. A miss biases the
\FNMR{} downward; a false removal costs one subject of resolution and
nothing else, since under a fixed model the surviving comparisons stay
valid. We therefore set the overlap cut at a low audit false-positive rate
and \emph{measure} the flags' precision by human verification
(\cref{sec:setup:human}); subject counts of cleaned sets are after
over-removal.

Deduplication links identity pairs whose cosine exceeds a threshold $T$ and
collapses each connected component (\cref{fig:circles}b). It has two
failure modes. \emph{1) Percolation}: if $T$ falls inside the bulk of the model's impostor
distribution, the linkage graph percolates into a giant component, a phase
transition rather than a population of duplicate people, so we report the
number of linked pairs and the largest component with every deduplication
(\cref{sec:protocol}). \emph{2) Censoring}: after removing every pair above
$T$ the remaining impostor maximum is at most $T$, so an operating point
whose threshold would have landed near $T$ is truncated and reads
optimistically; we print a point only when its threshold lies clearly below
$T$ (\cref{sec:protocol:noise}).

\subsection{Effective number of independent comparisons}
\label{sec:method:neff}

Write $Y_{ij}=\mathbf{1}[s_{ij}>\tau]$ for the indicator that impostor pair
$(i,j)$ falsely matches at threshold $\tau$, and $p=\mathbb{E}[Y_{ij}]$ for
the true \FMR{}. The estimate over $N$ pairs,
$\widehat{\FMR}=N^{-1}\sum_{(i,j)}Y_{ij}$, would have variance $p(1-p)/N$ if
the $Y_{ij}$ were independent. They are not: if subject $i$ takes part in
$k_i$ comparisons, the variance carries one covariance term for each of the
$\sum_i k_i(k_i-1)$ ordered pairs of comparisons sharing a subject,
\begin{equation}
\mathrm{Var}(\widehat{\FMR})=\frac{1}{N^2}\Big[N\,p(1-p)
  +\sum_{i}\sum_{j\ne l}\mathrm{Cov}(Y_{ij},Y_{il})\Big].
\label{eq:var}
\end{equation}
The \emph{effective number of independent comparisons} is the $N$ an
independent sample would need to reach the same variance,
\begin{equation}
\Neff=\frac{p(1-p)}{\mathrm{Var}(\widehat{\FMR})},
\label{eq:neff}
\end{equation}
and the number of independent false matches behind an operating point is
$\Neff\cdot\alpha$, which is what \cref{eq:floor} should be evaluated with.
Fogliato~\etal~\cite{fogliato2024ci} analyse \cref{eq:var} asymptotically for
1:1 matching: when the covariance term dominates,
$\mathrm{Var}(\widehat{\FMR})=O(1/G)$ in the number of subjects $G$, placing
$\Neff$ at a floor of order $G/2$; when the error rate approaches the
boundary the covariance shrinks and $\Neff$ moves toward $\binom{G}{2}$.
Their Wilson interval floors \cref{eq:neff} at $\lfloor G/2\rfloor$. Where a real
benchmark at \fmr{6} sits between $\binom{G}{2}$ and $G/2$ is therefore an
empirical question. We estimate $\mathrm{Var}(\widehat{\FMR})$ by
resampling subjects, estimate $\mathrm{Cov}(Y_{ij},Y_{il})$ over all
subject-sharing pairs, and report $\Neff$ with its own interval
(\cref{sec:neff}).

\section{Experimental Setup}
\label{sec:setup}

\subsection{Models}
\label{sec:setup:models}

We use four face-recognition models in two ways: an \emph{auditor} that performs the overlap search, and \emph{models under test} for which we report \FNMR{}. For all results in \cref{sec:audit,sec:masked}, the auditor is the publicly available ArcFace-R100~\cite{deng2019arcface,he2016resnet} trained on Glint360K~\cite{an2021glint360k} from the insightface model zoo\footnote{\url{https://github.com/deepinsight/insightface/tree/master/model_zoo}}. We validate the released ONNX weights by reproducing the published LFW accuracy ($99.82\%$). Glint360K extends MS1MV2~\cite{wu2024overlap}, and we treat it as disjoint from WebFace42M (\cref{sec:limitations}): the two may share individuals, but the audit requires only that the auditor was not trained to discriminate WebFace42M identities as classes.

The models under test are (i) our ViT-L and ViT-S models, trained within this codebase using the TransFace recipe~\cite{dan2023transface} on WebFace42M only (configurations \texttt{wf42m} ViT-L/ViT-S), and (ii) the public baseline \texttt{w600k\_r50}, an insightface ResNet-50~\cite{he2016resnet} trained on WebFace600K, a $600$K-identity subset of the WebFace corpus~\cite{zhu2021webface260m} (model-zoo package \texttt{buffalo\_l}). The baseline is never used as an auditor. Finally, the held-out protocol in \cref{sec:protocol} uses two auditors: the insightface ArcFace-R100 trained on MS1MV3~\cite{deng2019lightweight} (validated by reproducing the published LFW/CFP-FP/AgeDB-30 accuracies of $99.817/98.929/98.467$) and the Glint360K ArcFace-R100 described above.

\subsection{Corpora, probes, and conditions}
\label{sec:setup:corpora}\label{sec:setup:conditions}

The training corpus under audit is WebFace42M~\cite{zhu2023webface260m_pami}, obtained from the official distribution under its licence (\cref{sec:limitations}). We audit MS1MV3~\cite{deng2019lightweight}, IJB-C~\cite{maze2018ijbc}, LFW~\cite{huang2008lfw}, CFP-FP~\cite{sengupta2016cfp}, AgeDB-30~\cite{moschoglou2017agedb}, and the five masked-face datasets of \cref{sec:masked}, with probes of up to $c=8$ images per identity for IJB-C and the masked datasets and $c=2$ for MS1MV3 and the LFW family. For IJB-C we use the file-level subject groupings (which exceed the published subject count; see \cref{sec:neff}); for LFW, CFP-FP, and AgeDB-30 we use the groupings induced by each protocol's same-pair files, which can over-split individuals, so their overlap rates are lower bounds.

For the bias measurement, we sample up to four images per MS1MV3 identity uniformly at random ($371{,}934$ images) and score all pairs, treating within-identity pairs as genuine and cross-identity pairs as impostor. \cond{raw} denotes the unmodified MS1MV3 identities. \cond{clean} removes the identities flagged by the audit ($82{,}780$ at the calibrated threshold; three fewer than in \cref{fig:audit} due to rounding) and then deduplicates the remainder at $T=0.70$ (\cref{sec:method:cleaning}), yielding $10{,}102$ identities. \cond{random control} draws size-matched subsets uniformly at random from \cond{raw}, $20$ times, to isolate effects of subset size or difficulty. The $2\times2$ ablation adds \cond{remove only} ($10{,}651$ identities) and \cond{dedup only} ($91{,}212$). For each condition we set thresholds from that condition's impostor distribution (\cref{eq:operating_point}) and evaluate all three models under test. No model is retrained; we quantify reporting bias, not the causal effect of the overlapping identities~\cite{wu2024overlap}.

\subsection{Human verification: stratified sampling and reweighting}
\label{sec:setup:human}

We inspect the audit's flags on a sample stratified by flag cosine into the six bands of \cref{tab:human}, since precision is expected to be lowest near the threshold. A first batch samples $10$ subjects per band ($60$ sheets); a second adds $20$ to each of the two bands closest to the cut ($40$ sheets). Each sheet presents the matched pair together with three additional images from each side, shown in randomized order without the cosine value, band assignment, or identifier. If the training image contains no usable face, the sheet is marked \emph{unjudgeable} and excluded from the denominator.

Because sampling fractions differ by band, we estimate overall precision via population-weighted reweighting,
\begin{equation}
\widehat{\pi}=\frac{\sum_b \widehat{\pi}_b\, n_b}{\sum_b n_b},
\label{eq:reweight}
\end{equation}
where $\widehat{\pi}_b$ is the confirmed fraction among judged sheets in band $b$ and $n_b$ is the flagged population size of band $b$. The estimated net overlap is the product of $\widehat{\pi}$ and the flagged rate.

\subsection{Uncertainty}
\label{sec:setup:uncertainty}

All \FNMR{} confidence intervals are computed by bootstrapping at the identity level (rather than at the comparison level), recomputing \emph{both} the operating threshold and the \FNMR{} within each replicate. We use $B=2000$ bootstrap replicates and BCa intervals, with acceleration estimated via a leave-one-identity-out jackknife. The effective sample size analysis in \cref{sec:neff} is based on $1{,}000$ subject-level resamples of the IJB-C protocol under the official media-pooled template aggregation and detector-score weighting. For human verification, we compute Wilson intervals within each band and combine them using the reweighting in \cref{eq:reweight}.

\section{Results: Auditing Train--Test Identity Overlap}
\label{sec:audit}

\begin{figure}[t]
\centering
\includegraphics[width=\linewidth]{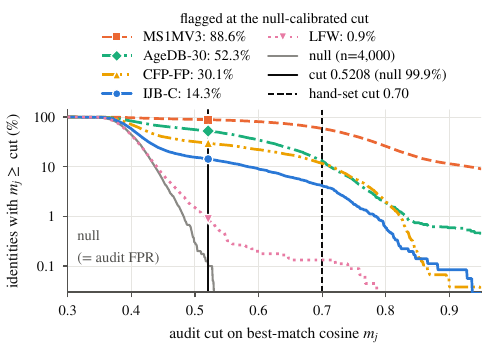}
\caption{Overlap audit against all $2{,}059{,}907$ WebFace42M
representatives (Glint360K auditor): percentage of identities whose
best-match cosine $m_j$ exceeds the cut, and of the self-masked null (audit
FPR). Markers: null-calibrated cut $0.5208$.}
\label{fig:audit}
\end{figure}

\paragraph{Null distribution and auditor agreement.}
Using the self-masked representatives described in \cref{sec:method:detector},
the Glint360K-based auditor yields a null distribution with median $0.3959$,
$95$th percentile $0.4511$, $99.9$th percentile $0.5208$, and maximum $0.5403$.
For the two WebFace42M-trained ViT-L variants of \cref{sec:setup:models}, the
corresponding null medians are lower: $0.3052$ (with $95$th percentile $0.3592$)
for the WebFace42M-only model and $0.3246$ for the private-corpus variant.
At their respective calibrated thresholds, the two ViT-L variants flag
$92.2\%$ and $92.0\%$ of MS1MV3 identities, compared with $88.6\%$ flagged by
the independent auditor (\cref{fig:audit}). Pairwise Jaccard agreement is
$\ge 0.958$: $82{,}592$ identities are flagged by all three auditors,
$86{,}337$ by at least one, and $398$ by exactly one. Throughout, we report
counts from the independent auditor, which are the most conservative.

\begin{figure*}[t]
\centering
\includegraphics[width=\textwidth]{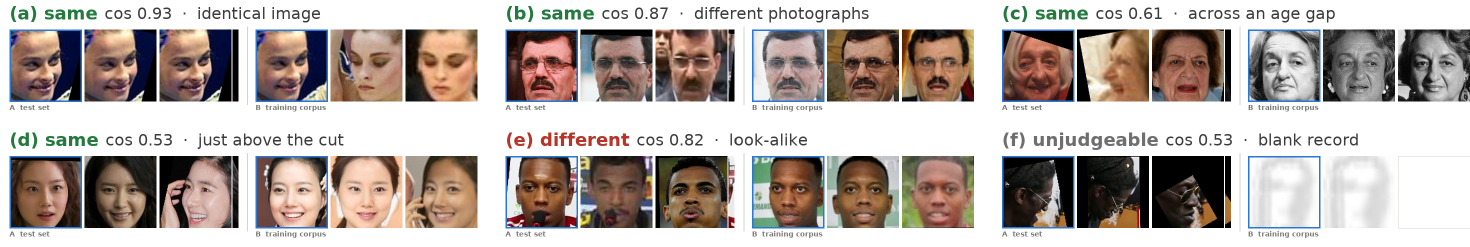}
\caption{Annotator sheets for six flagged IJB-C subjects (A: IJB-C; B:
WebFace42M; blue frame: pair attaining the cosine). (a)--(d): confirmed;
(e): look-alike; (f): blank training record.}
\label{fig:sheets}
\end{figure*}

\paragraph{MS1MV3 is $88.6\%$ contained in WebFace42M.}
At the calibrated cut, the audit flags $82{,}783$ of $93{,}431$ MS1MV3
identities and $510$ of $3{,}567$ IJB-C subjects.
Both corpora are assembled by scraping celebrity name lists; moreover,
WebFace260M is seeded from MS1M~\cite{zhu2023webface260m_pami}. Consequently,
MS1MV3 is not a suitable evaluation set for WebFace-trained models unless the
bias described in \cref{sec:bias} is explicitly removed.

\paragraph{IJB-C is retained despite being named in the de-overlap.}
The IJB-C flag count is largely insensitive to the threshold
(\cref{fig:audit}): $552$ subjects exceed cosine $0.50$, $322$ exceed $0.60$,
and $148$ exceed $0.70$ (the published removal threshold), with a maximum of
$0.9537$. In contrast, the null exhibits essentially no mass above $0.55$.
The published removal step removes $10{,}964$ identities
($0.53\%$)~\cite{zhu2023webface260m_pami}, and no cleaning code was released;
the procedure \emph{as reported} does not remove IJB-C from WebFace42M. The stated
threshold also conflicts with the merge rule in the same section
(\cref{sec:related:cleaning}), which removes training identities whose center
cosine lies between $0.5$ and $0.7$. Under the published description, the
$148$ subjects above $0.70$ should have been removed; the remaining $362$ lie
in $[0.52,0.70)$, a similarity band that the same discussion characterizes as
insufficiently reliable for training.

\paragraph{Named benchmarks are cleaned; unnamed benchmarks are not.}
LFW---the only verification set explicitly named in the de-overlap
statement---tracks the audit false-positive rate across thresholds
(\cref{fig:audit}). In contrast, AgeDB-30 and CFP-FP, which are reported in
the WebFace260M papers but not named in the de-overlap statement, exhibit
substantial flagged rates. This pattern is consistent with de-overlap having
been performed against a shorter benchmark list than the one ultimately used
for evaluation.

\subsection{Human verification of flagged matches}
\label{sec:audit:human}

\begin{table}[t]
\centering\footnotesize
\setlength{\tabcolsep}{3pt}
\resizebox{\linewidth}{!}{%
\begin{tabular}{lrrrrr}
\toprule
Cosine band & Flagged & Sampled & Judged & Confirmed & Unjudgeable \\
\midrule
$[0.52, 0.55)$ & 68  & 30 & 10 & $50.0\%$ & 20 \\
$[0.55, 0.60)$ & 120 & 30 & 13 & $69.2\%$ & 17 \\
$[0.60, 0.65)$ & 102 & 10 &  9 & $77.8\%$ &  1 \\
$[0.65, 0.70)$ & 72  & 10 &  7 & $71.4\%$ &  3 \\
$[0.70, 0.80)$ & 127 & 10 & 10 & $70.0\%$ &  0 \\
$[0.80, 1.00]$ & 21  & 10 &  9 & $77.8\%$ &  1 \\
\midrule
all            & 510 & 100 & 58 & $\mathbf{69.2\%}$ $[56.8, 81.7]$ & 42 \\
\bottomrule
\end{tabular}}%
\caption{Human verification of IJB-C flags (\cref{sec:setup:human}).
``Judged'': sheets with a usable training image; ``Confirmed'': fraction of
judged sheets showing the same person. The ``all'' row is
population-reweighted (\cref{eq:reweight}), with $95\%$ CI.}
\label{tab:human}
\end{table}

Among the judged sheets in \cref{tab:human}, $40$ depict the same person. Multiplying the
population-reweighted confirmation rate $\widehat{\pi}$ (``all'' row) by the
flagged rate yields the residual overlap remaining in IJB-C after the
published de-overlap procedure: $0.692\times14.3\%=9.9\%$, with interval
$[8.1,11.7]$.

Two observations from \cref{tab:human} are noteworthy. First, the confirmation
rate is approximately constant across cosine bands from $0.55$ to $1.00$ and
drops only in the band adjacent to the decision threshold; in particular, a
flag at $0.90$ is not materially more reliable than one at $0.60$. Second,
$37$ of the unjudgeable sheets fall in the two bands nearest the threshold.
Their training images are degenerate records (e.g., blank crops, uniform color
fields, severe blur; \cref{fig:sheets}(f)) that behave as similarity hubs: across
all flags, $34.5\%$ match to a representative with no discernible face,
compared with a $3.3\%$ base rate. Moreover, $30$ representatives account for
$184$ of the flags, indicating that many of the weakest flags are driven by a
small number of corrupted training records.

\section{Results: Bias at the Operating Point}
\label{sec:bias}

Having established the extent of train--test identity overlap, we now quantify its impact under the evaluation conditions described in \cref{sec:setup:conditions}.

\begin{figure}[t]
\centering
\includegraphics[width=\linewidth]{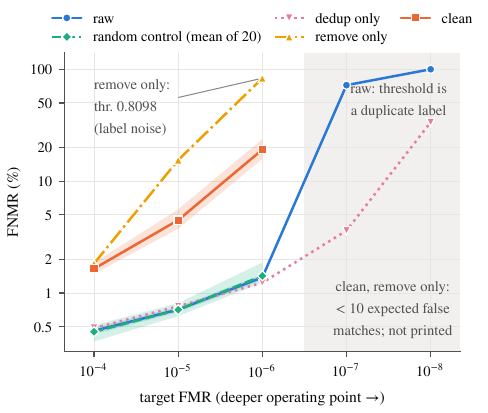}
\caption{\FNMR{} versus \FMR{} on MS1MV3 for a fixed model (our
WebFace42M ViT-L), varying only the test identities (\cref{sec:setup:conditions}).
Bands: identity-level BCa $95\%$ intervals; for the control, also the
min--max over $20$ draws.}
\label{fig:main}
\end{figure}

\subsection{Headline result}
\label{sec:bias:main}

\cref{fig:main} provides the central empirical result. At \fmr{4}, \fmr{5}, and \fmr{6}, \cond{raw} yields $0.46\%$, $0.71\%$, and $1.38\%$, whereas \cond{clean} yields $1.65\%$, $4.49\%$, and $19.14\%$. Excluding identities that overlap the training corpus increases \FNMR{} by $3.6\times$ at \fmr{4}, $6.3\times$ at \fmr{5}, and $13.9\times$ at \fmr{6}, indicating that the bias is substantial and varies with the operating point. In absolute terms, the gaps are $1.2$ percentage points at \fmr{4} and $17.8$ percentage points at \fmr{6}, substantially exceeding the biases reported by Wu~\etal~\cite{wu2024overlap} and by Dulhanty and Wong~\cite{dulhanty2020auditing} (\cref{sec:intro}).

\subsection{Ablation: random control and the \texorpdfstring{$2\times2$}{2x2} decomposition}
\label{sec:bias:ablation}

The three ablation conditions in \cref{fig:main} address two potential alternative explanations.

\paragraph{Is it subset size or difficulty?}
Because \cond{clean} is much smaller than \cond{raw}, one might suspect that the remaining identities are intrinsically more difficult. The \cond{random control} refutes this: at \fmr{6}, $20$ size-matched draws have standard deviation $0.14\%$ (range $[1.24,1.87]$) around a mean that is indistinguishable from \cond{raw} (\cref{fig:main}), whereas \cond{clean} lies more than $100$ control standard deviations away. Thus, subset size explains none of the effect.

\paragraph{Is it overlap removal or deduplication?}
The $2\times2$ design disentangles the two interventions. Deduplication alone (\cond{dedup only}) \emph{reduces} \FNMR{}@\fmr{6} from $1.38\%$ to $1.24\%$ (\cref{fig:main})
by removing duplicate pairs that would otherwise constrain the operating threshold. Applying overlap removal on top of deduplication then increases \FNMR{} by a factor of $15.4$. The bias is therefore due to overlap; label noise, if anything, attenuates it. The \cond{remove only} \fmr{6} value is an artefact: without deduplication its
threshold ($0.8098$) lies within the duplicate population, so it reflects annotation errors rather than the model (\cref{sec:method:cleaning}).

\paragraph{Does the \cond{clean} figure rest on noise?}
The \cond{clean} identities yield $\binom{G}{2}=5.10\times10^{7}$ independent impostor pairs, implying $51$ expected independent false matches
at \fmr{6}; we therefore do not report deeper operating points. With identity-level BCa resampling
(\cref{sec:setup:uncertainty}), the \cond{clean} \fmr{6} estimate of $19.14\%$ has a $95\%$ interval of $[15.8,23.8]$, compared with $[3.8,5.6]$ at
\fmr{5} and $[1.5,1.9]$ at \fmr{4}. The \cond{raw} \fmr{6} estimate of $1.38\%$ has interval $[1.30,1.46]$, and the corresponding inflation factor is $[11.5,17.2]$.

\subsection{The mechanism, measured}
\label{sec:bias:mechanism}

\"Ozt\"urk and Marcel~\cite{ozturk2026membership} show that seen and unseen identities differ in cluster geometry under margin-based losses. We measure how this shift reaches \FNMR{} via impostor-score quantiles, threshold displacement, and a counterfactual \FNMR{}.

For ViT-L, the seen and unseen impostor distributions are nearly identical through most of the mass: they agree to within $0.001$ cosine at the median ($0.0005$ vs. $0.0010$), at $p_{99}$ ($0.1075$ vs. $0.1080$), and at $p_{99.9}$ ($0.1490$ vs. $0.1500$).
The separation emerges only in the extreme tail: beyond $p_{99.99}$, the $1-10^{-6}$ quantiles are $0.3055$ (seen) and $0.5265$ (unseen). Overlap therefore truncates the tail (\cref{fig:densities}), which displaces the threshold: the \cond{raw} and \cond{clean} thresholds differ by $0.010$ at \fmr{4} ($0.1952$ vs. $0.2047$), by $0.079$ at
\fmr{5}, and by $0.188$ at \fmr{6} ($0.3386$ vs. $0.5267$).

\begin{figure}[t]
\centering
\includegraphics[width=\linewidth]{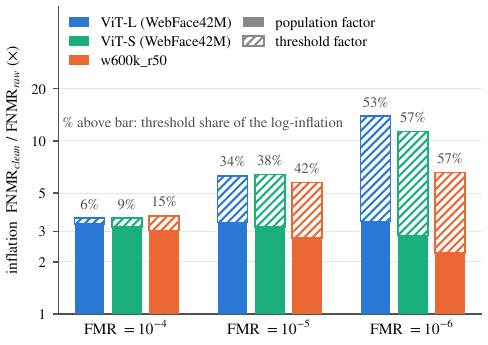}
\caption{\cond{clean}/\cond{raw} \FNMR{} ratio on MS1MV3 (log scale), split into a population factor (solid) and a threshold-displacement factor (hatched). Ratios at \fmr{4}, \fmr{5}, and \fmr{6}: $3.59$, $6.30$, $13.88$ (ViT-L); $3.58$, $6.41$, $11.34$ (ViT-S); $3.70$, $5.75$, $6.56$ (\texttt{w600k\_r50}).}
\label{fig:counterfactual}
\end{figure}

The third measurement decomposes the inflation into two components: a \emph{population} effect (unseen genuine pairs are harder even at a fixed threshold) and a \emph{threshold} effect (the additional error induced by the displaced threshold). Evaluating
\cond{clean} genuine pairs at the \cond{raw} threshold yields \FNMR{} values of $1.53\%$, $2.40\%$, and $4.72\%$ at \fmr{4}, \fmr{5}, and \fmr{6}, respectively, which lie between the \cond{raw} and \cond{clean} curves in \cref{fig:main}. The total inflation factors as \textit{population}$\times$\textit{threshold}; for ViT-L at \fmr{6},
$13.88=3.42\times4.06$ (\cref{fig:counterfactual}). The population factor is approximately constant with depth (about $3.4$ for ViT-L), whereas the
threshold factor increases with depth, accounting for $6\%$ of the log-inflation at \fmr{4}, $34\%$ at \fmr{5}, and $53\%$ at \fmr{6}; the other two models follow the same ordering.

\subsection{Across models: the margin halves}
\label{sec:bias:models}

Because the magnitude of inflation depends on the model's training corpus, it does not cancel when comparing models. Among the \fmr{6} ratios in \cref{fig:counterfactual}, the public \texttt{w600k\_r50} exhibits the smallest inflation factor. This should not be interpreted as superior performance: its \cond{raw} \FNMR{}@\fmr{6} is already $4.43\%$ (i.e., $3.2\times$ higher than ViT-L), and its \cond{clean} \FNMR{}@\fmr{6} remains the worst of the three at $29.06\%$. Its smaller factors are consistent with a $600$K-identity training subset having learned fewer of the flagged identities as classes; since the flags are defined relative to WebFace42M, this is an explanation rather than a measurement.

Across these three models, the ordering is preserved after cleaning, but the margin is not: the apparent $3.2\times$ advantage on \cond{raw} MS1MV3 at \fmr{6} shrinks to $1.5\times$ on \cond{clean}, implying that more than half of the original gap is attributable to corpus contamination. On DigiFace-1M~\cite{bae2023digiface} (\cref{tab:digiface}), which is synthetic and therefore has zero overlap by construction, \texttt{w600k\_r50} leads at \fmr{4} whereas our ViT-L leads at \fmr{8}.

\begin{table}[t]
\centering\footnotesize
\setlength{\tabcolsep}{4pt}
\begin{tabular}{lrrrrr}
\toprule
Model & $10^{-4}$ & $10^{-5}$ & $10^{-6}$ & $10^{-7}$ & $10^{-8}$ \\
\midrule
ViT-L (ours)        & $11.45$ & $21.53$ & $34.17$ & $48.13$ & $\mathbf{63.29}$ \\
ViT-S (ours)        & $12.89$ & $24.04$ & $38.29$ & $55.56$ & $83.76$ \\
\texttt{w600k\_r50} & $10.38$ & $20.54$ & $34.53$ & $51.39$ & $71.97$ \\
\bottomrule
\end{tabular}
\caption{\FNMR{} (\%) on DigiFace-1M ($99{,}999$ synthetic identities, five images each; zero overlap), image-to-image without template aggregation.}
\label{tab:digiface}
\end{table}

\section{Results: How Many Independent Impostor Comparisons Does IJB-C Have?}
\label{sec:neff}

\begin{figure}[t]
\centering
\includegraphics[width=\linewidth]{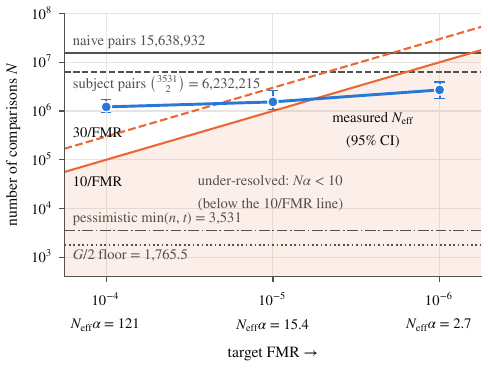}
\caption{$\Neff$ for IJB-C's official protocol ($95\%$ CI) against the three counting conventions of \cref{sec:method:resolution} and the $G/2$ floor~\cite{fogliato2024ci}. Below the axis: expected independent false matches, $\Neff\alpha$.}
\label{fig:neff}
\end{figure}

We now estimate $\Neff$ (\cref{eq:neff}) directly for IJB-C's official 1:1 protocol~\cite{maze2018ijbc}, which specifies $19{,}557$ mated comparisons and $15{,}638{,}932$ non-mated template comparisons over $23{,}124$ templates.
Union--find on the mated pairs yields $3{,}567$ connected components; $36$ are isolated templates and the remainder matches the published subject count.
The three counting conventions of \cref{sec:method:resolution} imply $15{,}638{,}932$, $6{,}232{,}215$, and $3{,}531$ impostor comparisons, a factor of $4{,}400$ apart.
\cref{tab:neff} and \cref{fig:neff} report the estimate at the thresholds that realise each \FMR{} for Glint360K ArcFace-R100 (\cref{sec:setup:uncertainty}).

\begin{table}[t]
\centering\footnotesize
\setlength{\tabcolsep}{4pt}
\begin{tabular}{lccc}
\toprule
& \fmr{4} & \fmr{5} & \fmr{6} \\
\midrule
$\Neff$ ($10^{6}$)        & $1.21$ & $1.54$ & $2.70$ \\
\quad $95\%$ CI          & $[0.93,1.74]$ & $[1.08,2.65]$ & $[1.78,3.92]$ \\
observed false matches   & $1{,}563$ & $156$ & $15$ \\
$\mathrm{Cov}(Y_{ij},Y_{il})$ & $4.6{\cdot}10^{-8}$ & $3.3{\cdot}10^{-9}$ & $1.1{\cdot}10^{-10}$ \\
$\Neff$ / naive count    & $8\%$ & -- & $17\%$ \\
\midrule
rule of ten / rule of 30  & yes / yes & yes / no & no / no \\
\bottomrule
\end{tabular}
\caption{Effective number of independent impostor comparisons in IJB-C's official protocol (Glint360K ArcFace-R100; $1{,}000$ subject-level resamples).}
\label{tab:neff}
\end{table}

None of the three conventions matches the estimate: the confidence interval excludes all three at every operating point.
$\Neff$ is approximately one fifth of the subject-pair count at \fmr{4} and below one half at \fmr{6}, yet roughly three orders of magnitude above the $G/2$ floor.
As Fogliato~\etal{} predict, the within-subject covariance falls with depth, so impostor comparisons become more nearly independent where false matches are rare.
This does not justify reporting \fmr{6}: reportability is governed by $\Neff\cdot\alpha$, whose interval $[1.8,3.9]$ at \fmr{6} satisfies neither rule.
IJB-C's official protocol therefore supports \fmr{5} under common practice and \fmr{4} under biometrics standards, but \fmr{6} under neither.
This is close to the minimum reportable \FMR{} of $6\times10^{-5}$ of Fallahi~\etal~\cite{fallahi2025reliability}, obtained by setting aside subject dependence; estimating $\Neff$ replaces that assumption with a measurement.

\section{Masked-Face Benchmarks}
\label{sec:masked}

Masked-face benchmarks were largely released after the major training corpora were assembled. We apply the audit of \cref{sec:audit} to five of them, probing with their \emph{bare} (unmasked) images, since masked probes would obscure overlap.

\begin{table}[t]
\centering\footnotesize
\setlength{\tabcolsep}{3pt}
\resizebox{\linewidth}{!}{%
\begin{tabular}{lrrrrl}
\toprule
Benchmark & Identities & median $m$ & flagged & audit FPR & origin \\
\midrule
RMFD~\cite{wang2023masked}         & 442     & 0.6107 & \textbf{265 (60.0\%)} & $0.1\%$ & web, CN celebrities \\
RMFVD~\cite{wang2023masked}        & 426     & 0.5135 & \textbf{210 (49.3\%)} & $0.1\%$ & web, CN celebrities \\
MDMFR~\cite{ullah2022mdmfr}        & 226     & 0.3939 & 0 (0.0\%)             & $0.1\%$ & collected, PK \\
COMASK20~\cite{nguyen2021comask}   & 278     & 0.3947 & 0 (0.0\%)             & $0.1\%$ & collected, VN \\
PKU-Masked-Face~\cite{ding2020pku} & 1{,}018 & 0.4214 & 3 (0.3\%)             & $0.1\%$ & collected, lab \\
\bottomrule
\end{tabular}}
\caption{Audit results (\cref{fig:audit}) for five masked-face benchmarks, using the same auditor and decision threshold ($0.5208$). ``Origin'' indicates whether identities were scraped from a celebrity name list or collected via controlled photography.}
\label{tab:audit_masked}
\end{table}

\paragraph{Provenance determines overlap.}
RMFD and RMFVD, the two most commonly used masked-face benchmarks, are scraped from Chinese celebrity name lists and have $60.0\%$ and $49.3\%$ of identities flagged (\cref{tab:audit_masked}). The three collected datasets are at or near the audit's false-positive rate. Provenance, rather than size or release date, determines overlap.

\paragraph{Resolution after cleaning.}
After removing flagged identities, RMFD retains $161$ identities with both bare and masked imagery, yielding $12{,}880$ independent subject pairs and a resolution floor (\cref{eq:floor}) of $7.8\times10^{-4}$. Consequently, the cleaned RMFD supports operating points down to \fmr{3}, but not beyond.

Under the same filtering criterion, PKU-Masked-Face retains $1{,}015$ identities, corresponding to $514{,}605$ independent subject pairs and a floor of $1.9\times10^{-5}$. A masked-face benchmark that resolves \fmr{4} with verified absence of overlap therefore exists, but it is not the one the literature emphasises.

\paragraph{Pooling across benchmarks.}
Under the same threshold, RMFD and RMFVD link $358$ identities to $352$ across $400$ links, so RMFVD largely re-curates the same individuals and their subject counts cannot be summed. MDMFR, COMASK20, and PKU-Masked-Face link neither to one another nor to the scraped sets, so only they can be pooled.

\section{Towards a Usable Deep-Tail Protocol}
\label{sec:protocol}

Existing real-face benchmarks either overlap major training corpora (\cref{sec:audit}) or are too small for the operating points at which they are reported (\cref{sec:neff}). This section turns these constraints into a deep-tail protocol.

\subsection{Label noise sets the floor}
\label{sec:protocol:noise}

Deep thresholds are set by the upper tail of the impostor distribution (\cref{eq:operating_point}), the region most sensitive to label noise. On \cond{raw} MS1MV3 the maximum impostor score is $1.0000$, caused by the same image appearing under two labels; it pushes the \fmr{7} threshold to $0.7920$, and at \fmr{8} the threshold coincides with the duplicate (\cref{tab:noise}). \cref{tab:noise} also quantifies the two failure modes of \cref{sec:method:cleaning}. \emph{Percolation}: a cut at $0.4$ would lie in the bulk of ViT-L's impostor distribution (its \cond{raw} \fmr{6} threshold is $0.3386$ and the audit null median $0.3052$), whereas at $T=0.70$ the linkage graphs have small largest components, consistent with small populations of duplicates. \emph{Censoring}: the \cond{clean} \fmr{6} threshold ($0.5267$) lies well below $T$, but the \cond{dedup only} \fmr{8} threshold ($0.6735$) approaches it, the regime that the tooling marks as censored.

\begin{table}[t]
\centering\footnotesize
\setlength{\tabcolsep}{3pt}
\resizebox{\linewidth}{!}{%
\begin{tabular}{lrrrr}
\toprule
Condition & \FNMR{}@$10^{-7}$ & \FNMR{}@$10^{-8}$ & linked pairs & largest comp. \\
\midrule
\cond{raw}        & $72.11$ & $99.99$ & -- & -- \\
\cond{dedup only} & $3.61$  & $33.54$ & $2{,}357$ & $7$ \\
\cond{clean}      & --      & --      & $603$ & $6$ \\
\bottomrule
\end{tabular}}
\caption{Label noise on MS1MV3 (ViT-L): \FNMR{} (\%) before and after deduplication at $T=0.70$, with the size of the linkage graph.}
\label{tab:noise}
\end{table}

\subsection{What the available corpora can support}
\label{sec:protocol:corpora}

\begin{figure}[t]
\centering
\includegraphics[width=\linewidth]{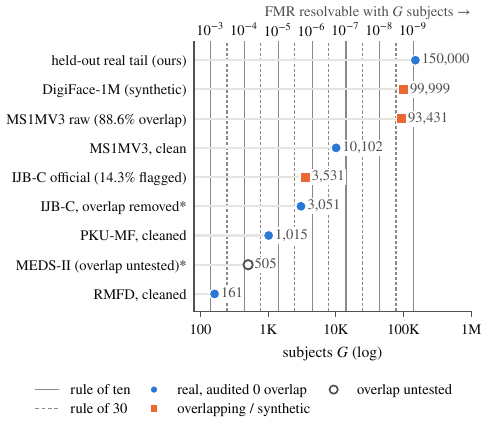}
\caption{Deep-tail capability of public evaluation corpora: subject counts required per operating point under the subject-pair count (\cref{eq:subject_pairs}; solid: rule of ten, dashed: rule of 30). A corpus resolves the curves to its left. Filled circles: real-face sets with audited zero overlap; squares: overlapping or synthetic; open circle: MEDS-II~\cite{founds2011meds} (off-domain, not audited). $^*$Counts not re-derived from the released result files.}
\label{fig:capability}
\end{figure}

\cref{fig:capability} aggregates the relevant subject counts. MEDS-II ($505$ subjects) resolves \fmr{4}; IJB-C resolves \fmr{5} with $3{,}531$ subjects or the $3{,}051$ remaining after overlap removal; \cond{clean} MS1MV3 resolves \fmr{6}. No published real-face benchmark without train overlap supports deeper operating points; the only corpus with sufficient scale and zero overlap by construction is synthetic (DigiFace-1M; $99{,}999$ identities).

\subsection{Closing the gap by inverting the construction}
\label{sec:protocol:inversion}

Holding identities out of a fixed model is a data-acquisition problem. Instead, we fix public models with known training corpora, ArcFace-R100 trained on MS1MV3 and on Glint360K (\cref{sec:setup:models}), and subtract their training identities from a large public corpus: we audit WebFace42M against MS1MV3 (\cref{sec:method:detector}) and remove the \emph{union} of the flags of all three auditors (over-removal is acceptable; \cref{sec:method:cleaning}). \cref{tab:pool} gives the resulting pool; the benchmark samples $150{,}000$ of its identities, whose $1.12\times10^{10}$ subject pairs support \fmr{8} with an elevenfold margin over the rule of ten.

\begin{table}[t]
\centering\footnotesize
\setlength{\tabcolsep}{4pt}
\resizebox{\linewidth}{!}{%
\begin{tabular}{lr}
\toprule
WebFace42M identities removed (union)            & $85{,}014$ ($4.1\%$) \\
remaining                                         & $1{,}974{,}893$ \\
with more than eight images (median $19$)         & $1{,}148{,}862$ \\
deep-tail pool (minus $20{,}000$ for adaptation)  & $1{,}128{,}862$ \\
\quad independent subject pairs / floor           & $6.37\times10^{11}$ / $1.6\times10^{-11}$ \\
benchmark sample / subject pairs                  & $150{,}000$ / $1.12\times10^{10}$ \\
\bottomrule
\end{tabular}}
\caption{Construction of the held-out real-face pool.}
\label{tab:pool}
\end{table}

\begin{table}[t]
\centering\footnotesize
\setlength{\tabcolsep}{3pt}
\resizebox{\linewidth}{!}{%
\begin{tabular}{lrrrrrr}
\toprule
Model & $10^{-4}$ & $10^{-5}$ & $10^{-6}$ & $10^{-7}$ & $10^{-8}$ & pairs $>0.70$ \\
\midrule
R100 / MS1MV3     & 6.55 & 11.69 & 18.52 & 27.71 & 45.36 & $12$ \\
R100 / Glint360K$^\dagger$ & 4.29 & 9.26 & 23.79 & 61.90 & 91.86 & $533$ \\
DigiFace-1M$^\ddagger$, R100 / MS1MV3 & 34.86 & 51.08 & 66.12 & 78.36 & 87.44 & -- \\
\midrule
\multicolumn{7}{l}{\emph{the same protocol, sampling each identity's first images}} \\
R100 / MS1MV3     & 0.15 & 0.38 & 0.85 & 1.71 & 3.16 & -- \\
\bottomrule
\end{tabular}}
\caption{\FNMR{} (\%) on the held-out deep-tail protocol ($150{,}000$ subjects, four random images each; one impostor score per subject pair, the maximum of its $16$ comparisons; floor $8.9\times10^{-10}$; intervals $\le\pm0.1$ points). Last column: subject pairs above $0.70$ (percolation check). $^\dagger$Held out of MS1MV3, not of Glint360K. $^\ddagger$$98{,}770$ subjects, $4.88\times10^{9}$ pairs, floor $2.1\times10^{-9}$.}
\label{tab:realtail}\label{tab:percolation}
\end{table}

\cref{tab:realtail} supports four conclusions.
\textbf{First, the protocol resolves \fmr{7} and \fmr{8} on real faces with verified absence of overlap}; the MS1MV3-trained model is the cleanest case because its training identities are exactly those subtracted.
\textbf{Second, image sampling must be specified.} Selecting each identity's \emph{first} four records instead of a random four lowers \FNMR{} at \fmr{6} by $22\times$ (last row), although thresholds differ by only $0.009$ ($0.4564$ vs.\ $0.4651$): the first images of a scraped identity are more homogeneous (mean genuine cosine $0.772$ vs.\ $0.601$).
\textbf{Third, a pool is held out of a particular corpus, not of models in general.} The Glint360K-trained model yields $44\times$ more pairs above $0.70$, consistent with residual duplication within WebFace42M that it recognises, which also explains why it is better at \fmr{4} yet worse at \fmr{8}.
\textbf{Fourth, synthetic data is no substitute}: the same model has $3.6\times$ higher \FNMR{} on DigiFace-1M than on held-out real faces at \fmr{6}.
Because the protocol requires models \emph{not} trained on WebFace, it is not a leaderboard but a construction that any group with a public corpus and a public checkpoint can instantiate.

\subsection{Recommendations}
\label{sec:protocol:recommendations}

(1)~Report subject count and independent subject-pair count alongside every \FMR{} operating point, and do not report operating points that cannot be resolved by the available subjects (\cref{fig:capability}).
(2)~Audit the test set against the training corpus using a matcher trained on neither, calibrate the decision threshold using a masked-self null, and report the residual overlap rate in addition to the number of identities removed.
(3)~Deduplicate at a threshold that lies in the tail of the \emph{specific} model's impostor-score distribution; report the largest connected component as a percolation diagnostic; and mark operating points censored by cleaning.
(4)~Report uncertainty intervals obtained by resampling subjects rather than comparisons.

The audit, deduplication, deep-tail scoring, and resolution-checking code will be released publicly.

\section{Limitations}
\label{sec:limitations}

\paragraph{The audit provides a lower bound and remains model-dependent.}
We use a single representative image per training identity; using additional
images would necessarily increase the number of flagged overlaps. The
null-calibrated cutoff avoids the arbitrariness of a manually chosen
threshold, but it does not remove dependence on the underlying embedding.
The three agreeing matchers (\cref{sec:audit}) are all margin-trained. Likewise, we assume---rather than
directly measure---that Glint360K and WebFace42M are disjoint
(\cref{sec:setup:models}). A third-corpus audit of both datasets would
replace this assumption with an empirical estimate.

\paragraph{One training corpus and no retraining.}
All bias measurements are computed with respect to WebFace42M. Whether the
same operating-point dependence holds for models trained on Glint360K or
MS1MV2 remains untested, although the proposed mechanism relies only on
margin-based training. Moreover, our analysis targets \emph{reporting} bias,
not the causal effects studied by Wu~\etal~\cite{wu2024overlap}.

\paragraph{The headline comparison is limited by the number of subjects.}
The \cond{clean} subject count carries \fmr{6} beyond the rule of ten but
not the rule of 30 (\cref{sec:bias:ablation}), so the \fmr{6} factor has a
wide interval; the robust conclusions are the \emph{ordering} of conditions
and the \emph{increase with depth}.

\paragraph{Human verification is sampled rather than exhaustive.}
In contrast to the exhaustive annotation of Wu~\etal~\cite{wu2024overlap}, we
annotate a single sample with a single annotator, and no score band is
supported by more than a small number of judged sheets (\cref{tab:human}).
These rates therefore reflect the
order of magnitude of the audit's precision. The treatment of unjudgeable
sheets is also not uniquely determined: we exclude them because the
annotator makes no claim. If all unjudgeable sheets are counted as false
flags, the rate is $50.7\%$ $[39.7,61.6]$; if all are counted as true flags,
it is $79.5\%$ $[69.6,89.5]$.

\paragraph{Thresholds are alignment-specific.}
All thresholds are computed under a single alignment configuration and do
not transfer to a different detector or crop size.

\paragraph{The audited corpus is no longer distributed.}
We obtained WebFace42M from the official site under its licence in December
2024. The site now resolves to a registrar parking page, and the Internet
Archive records the download page as failing by 2025-06-20. Our copy cannot
be re-obtained for independent verification, which motivates
\cref{sec:protocol}.

\paragraph{Ethics.}
All corpora used here are
public research datasets accessed under their licences and, to our
knowledge, have not been withdrawn. Because the audit could implicitly
identify individuals who appear in both a training corpus and a benchmark---
a linkage to which they did not consent---we report only aggregate rates and
publish the sheets in \cref{fig:sheets} derived from the benchmarks' own
published images, but we do not release identity lists.

\section{Conclusion}
\label{sec:conclusion}

Train--test identity overlap is not a small additive bias. At the false
match rates where face recognition is deployed it inflates reported false
non-match rates by $3.6\times$ at \fmr{4}, $6.3\times$ at \fmr{5} and
$13.9\times$ at \fmr{6} on MS1MV3 under a fixed WebFace42M-trained model
(\cref{fig:main}); the inflation grows with
operating depth, so no constant correction exists, and a same-size random
control shows none of it. The largest public training corpus publishes a
de-overlap procedure that, as described, leaves $14.3\%$ of IJB-C subjects
flagged as present, $69.2\%$ of them confirmed by eye, at a threshold
stricter than the one the same section applies to its own training
identities; and the benchmark on which \fmr{6} figures are routinely
published has $2.7$ expected independent false matches at that point,
fewer than any accepted reporting rule requires (\cref{fig:neff}).

The cheap remedy is available today: report the subject count beside every
operating point and refuse to print the ones it cannot support. The
constructive remedy inverts the framing: holding identities out of a model
is an acquisition problem only while the model is fixed, and fixing the
corpus instead yields a real-face protocol that resolves \fmr{8} with
audited zero overlap, though it cannot score WebFace-trained systems.

{
    \small
    \bibliographystyle{ieeenat_fullname}
    \bibliography{main}
}

\end{document}